# Now that I Have a Good Theory of Uncertainty, What Else Do I Need? *

Piero P. Bonissone
Artificial Intelligence Program
General Electric Corporate Research and Development
Schenectady, New York 12301
Arpanet: bonissone@crd.ge.com

## Abstract

Rather than discussing the isolated merits of a normative theory of uncertainty, this paper focuses on a class of problems, referred to as Dynamic Classification Problem (DCP), which requires the integration of many theories, including a prescriptive theory of uncertainty.

We start by analyzing the Dynamic Classification Problem (DCP) and defining its induced requirements on a supporting (plausible) reasoning system. We provide a summary of the underlying theory (based on the semantics of many-valed logics) and illustrate the constraints imposed upon it to ensure the modularity and computational performance required by the applications.

We describe the technologies used for knowledge engineering (such as object-based simulator to exercise requirements, and development tools to build the KB and functionally validate it). We emphasize the difference between development environment and run-time system, describe the rule cross-compiler, and the real-time inference engine with meta-reasoning capabilities.

Finally, we illustrate how our proposed technology satisfies the DCP's requirements and analyze some of the lessons learned from its applications to situation assessment problems for Pilot's Associate and Submarine Commander Associate.

## 1 Normative vs. Prescriptive Theories of Uncertainty

The search for a *normative* uncertainty theory to be used in reasoning systems has been a major driving force in our research community. In the past, we have witnessed long discussions [Che85] defending the rationale of Cox's probability axioms [Cox46]. We have also argued about the merits of many proposed revisions: the proponents of interval-based representations have questioned Cox's axiom related to the sufficiency of a single number to represent uncertainty; the proponents of possibility measures have provided reasonable modifications to Cox's first and third axioms and derived from them solutions that are not probability measures [DP88]; others have argued that Cox's second axiom is only natural to a reduced set of people (familiar with conditional probabilities) [Sha88]. More recently, these religious wars have subsided, and a slightly more tolerant view has emerged. Uncertainty tools have been divided into *extensional* and *intensional* approaches, according to their respective focus on computational efficiency or purer semantics [Pea88]. In the 1988 Uncertainty Workshop, there has been an increased awareness of classes of problems requiring a *prescriptive* rather than a normative approach to reasoning with uncertainty.

Therefore, instead of proposing a modified uncertainty theory or a new application, we will describe one of such classes of problems and illustrate the complex evolution required for its solution. We will take the reader from the conception of a theory of uncertainty, through the development of an embedding technology (subject to the constraints induced from integrating it with other complementary theories), to the development and deployment of a knowledge-based system for one of many instances of such problem class. In the next section, we will analyze the Dynamic Classification Problem (DCP), define its induced requirements on a supporting (plausible) reasoning system, and illustrate some of technical risks involved in applying such reasoning system. This will be followed, in the third section, by a brief description of the underlying theory and of the constraints followed to implement its embedding technology. In

*This work was partially supported by the Defense Advanced Research Projects Agency (DARPA) under USAF/Rome Air Development Center contract F30602-85-C-0033. Views and conclusions contained in this paper are those of the authors and should not be interpreted as representing the official opinion or policy of DARPA or the U.S. Government.

the fourth section, we will describe the technologies used for knowledge engineering, emphasize the difference between development environment and run-time system, and explain their control of reasoning. Finally, in the fifth section, we will illustrate how our proposed technology satisfies the DCP's requirements and we will analyze some of the lessons learned from its applications.

## 2  Dynamic Classification Problems: Situation Assessment

The Classification Problem (CP) has been first introduced by Clancey [Cla84] in 1984. CP consists of recognizing a situation from a collection of data and selecting the best action in accordance with some objectives. The classification problem has a recurrent solution structure: a collection of data, generated from several sources, is interpreted as a predefined pattern. The recognized pattern is mapped into a set of possible solutions, from which one is selected as the most appropriate for the given case. This process is considered a *static* classification problem, since the information is assumed to be invariant over time or at least invariant over the time required to obtain the solution.

A more challenging case is the *Dynamic Classification Problem* (DCP), originally described in [BW88], in which the environment from which data are collected changes at a rate comparable with the time required to obtain a refined solution. Examples of such *dynamic* classification problems are real-time situation assessment (e.g., air traffic control), real-time process diagnosis (e.g., airborne aircraft engine diagnosis), real-time planning, and real-time catalog selection (e.g., investment selection during market fluctuations). The characteristic structure of this class of dynamic classification problems is illustrated in Figure 1.

Situation assessment (SA) [SBBG86], as part of the more extensive battlefield management problem, is a prototypical case of the dynamic classification problem. The retrospective component of situation assessment consists of analyzing and associating observed events to identify and understand those which are relevant. The prospective component consists of projecting the relevant events and assessing their future impact. The correct assessment of current and future impacts of a given situation requires continuous classification in a dynamic environment.

The output of the SA module feeds a Tactical Planner (TP) module, usually based on reactive planning techniques, which generates a sequence of corrective actions to be implemented in the dynamic world.

### 2.1  Requirements for the Reasoning Tool

The development of reasoning systems addressing SA problems is characterized by a variety of requirements, and an associated set of risks. We have classified these risks into the following three categories that will be brefly described: *inadequate knowledge representation, inference, and control; unwanted side effects of the methodology used in the development; classical Software Engineering problems applied to AI software*. In Sections four and five, we will illustrate the approaches and aids developed with our technology to overcome such risks.

#### 2.1.1  Inadequate Knowledge Representation, Inference, and Control

The first type of risk can arise from the inadequate selection of the reasoning tool used to develop the SA application. Due to the characteristics of the dynamic classification problems, such reasoning tool must be able to:

1. Efficiently and correctly represent and propagate uncertain information through the knowledge base. Such uncertainty representation and calculus must be modular enough to support dynamic changes in the boundaries of the available knowledge base, which might be dictated by the control mechanism.

2. Represent defaults (i.e., decision maker's preferences or biases) to be used as surrogates for missing information. Such defaults must be selected from a set of competing extensions to the current inference base. The extension selection must be maximally (or at least sufficiently) compatible with the available uncertain knowledge.

3. Retract conclusions, based on defaults or uncertain information, when such conclusions are inconsistent with new reliable evidence. This requirement requires an integration of the uncertain-belief revision system with the default reasoning system.

4. Incrementally update the knowledge base to avoid unnecessary computations. This efficiency requirement usually imposes constraints upon the type of network (usually a DAG) that must be maintained by the uncertain-belief revision system.

5. Maintain an awareness of time requirements and resources availability to guarantee a real-time response. This performance requirement usually causes the need of a meta-reasoning system (resource controller). Such controller, which is itself part of the system's overhead, will use compile-time knowledge about rule execution-time and bounds on the amount of information content provided to determine, at run-time, the best (or the satisficing) answer within the time constraints.

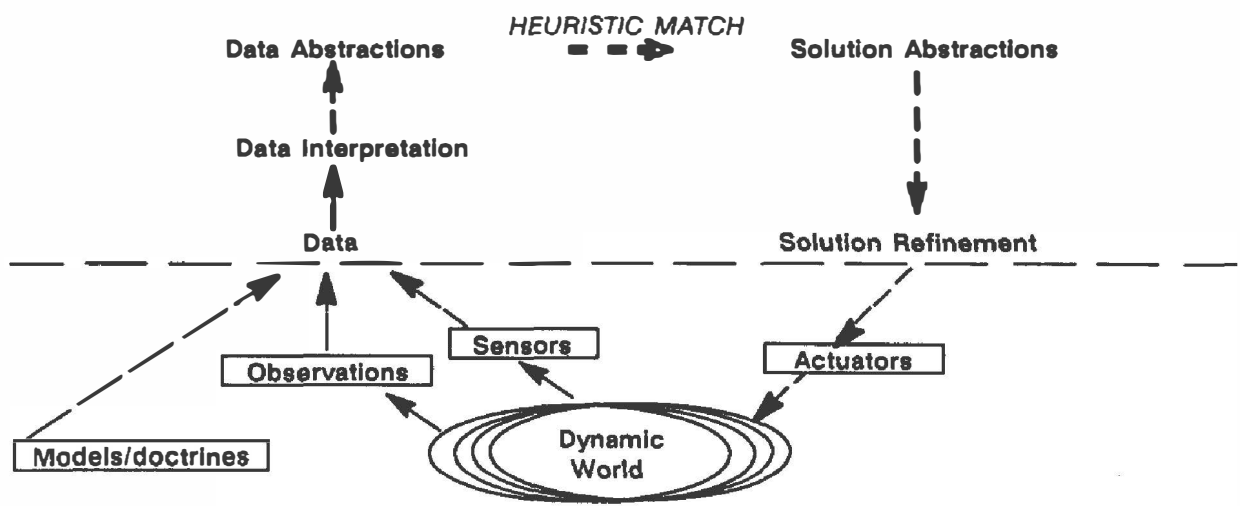

Figure 1: The Dynamic Classification Problem

#### 2.1.2 Unwanted Side Effects of the Methodology Used in the Development

Due to the evolving requirements of the SA class of problems, these applications undergo a large number of iterative refinements, as described by the rapid prototyping paradigm [Pre87]. This constant feedback allows the Knowledge Base architect to develop a working understanding of the problem, of the interactions between the various modules into which the original problem has been functionally decomposed, of the message traffic between these modules, etc. This information is then used by the KB architect to reassign functions to modules, to modify data structures, to establish new communication lines, to determine global vs. local variables, etc. However, rapid prototyping may induce a sloppy KB designer into the fallacy that an original *toy* problem and its corresponding design will easily scale-up to a full-size application. Thus, the ease of scalability from small prototypes is a constant requirement to be addressed in this type of problems.

#### 2.1.3 Classical Software Engineering Problems Applied to AI Software

The development of Knowledge-based systems presents numerous difficulties common to the development of more conventional software. The presence of uncertainty in the reasoning process further complicates these issues. The most common problems are: *KB functional validation*, i.e. testing the correctness and completeness of the rule set; *KB performance validation*, i.e., verifying that the application's response-time meets the timing requirements; *portability of application to different platforms*, i.e., providing a technology transfer path that enables multiple run-time versions, while still ensuring software maintenance. In Section 4 we will illustrate our proposed solutions to these requirements.

## 3 RUM's Theory and Constraints

RUM (Reasoning with Uncertainty Module) is a system for reasoning with uncertainty whose underlying theory is anchored on the semantics of many-valued logics [Bon87b]. Such system provides a representation layer to capture structural and numerical information about the uncertainty, an inference layer to provide a selection of truth-functional triangular-norm based calculi [BD86], and a control layer to focus the reasoning on subset of the KB, to (procedurally) resolve ignorance and conflict, and to maintain the integrity of the inference base via a belief revision system.

### 3.1 Triangular Norms and Multi-Valued Logics

RUM's inference layer is built on a set of five Triangular norms (T-norms) based calculi. The theory of T-norms and its underlying calculi have been covered in previous articles [Bon87b], [BD86], [Bon87a], [Bon87c], [BGD87]. This subsection is included for the reader's convenience.

Triangular norms (T-norms) and Triangular conorms (T-conorms) are the most general families of binary functions that satisfy the requirements of the conjunction and disjunction operators, respectively. T-norms and T-conorms are two-place functions from [0,1]x[0,1] to [0,1] that are monotonic, commutative and associative. Their corresponding boundary conditions, i.e., the evaluation of the T-norms and T-conorms at the extremes of the [0,1] interval, satisfy the truth tables of the logical AND and OR operators.[1]

The five T-norms used in RUM are:

$$T_1(a,b) = max(0, a+b-1)$$

$$T_{1.5}(a,b) = (a^{0.5} + b^{0.5} - 1)^2 \quad \text{if } (a^{0.5} + b^{0.5}) \geq 1$$
$$\qquad\qquad = 0 \quad \text{otherwise}$$

$$T_2(a,b) = ab$$

$$T_{2.5}(a,b) = (a^{-1} + b^{-1} - 1)^{-1}$$

$$T_3(a,b) = min(a,b)$$

Their corresponding DeMorgan dual T-conorms, denoted by $S_i(a,b)$, is defined as:

$$S_i(a,b) = 1 - T_i(1-a, 1-b)$$

All the calculi operations required in RUM can be expressed as a function of a T-norm and a negation operator. These operations are further elaborated upon in [Bon87b].

### 3.2 RUM's Constraints: Descriptions and Definitions

The decision procedure for a logic based on real-valued truth values is (potentially) much more computationally expensive than the decision procedure for crisp logic. In crisp logic only one proof is needed to establish the validity of a theorem. In real-valued logic all possible proofs must be explored in order to ensure that the certainty of a proposition has been maximized. RUM deals with the possible computational explosion through a series of trade-offs:

1. RUM allows the user to create rules with variables (rule-templates). These variables represent complex objects with multiple attributes. The typical problems of first order reasoning are avoided by instantiating the rule templates at run time. The implicit universal quantifier (that determines the scope of the variables in the rules) is replaced by on ongoing enumeration of the instances of the objects encountered by the reasoning system. Thus, a single rule may give rise to many rule instances at run time, but all these rule instances will be propositional.

2. RUM does not allow cyclic (monotonic) rules. Given the rule $P \Rightarrow_n^s Q$, RUM uses the amount of confirmation of $P$, the degree of sufficiency $s$, and *modus ponens* to derive the amount of confirmation of $Q$. Similarly, by using the amount of refutation of $P$, the degree of necessity $n$ and *modus tollens*, RUM derives the amount of refutation of $Q$. The other two modalities (necessity and *modus ponens*, sufficiency and *modus tollens*) are not used, as they would determine values of P from Q.

   Note that most T-norms are strictly monotonically decreasing. Cyclic rules that propagate their truth values using these T-norms will continually cycle until all the truth values are 0 (unless relaxations, tagging, or other methods were used). Only one T-norm (*min*) satisfies the axioms of a (pseudo-complemented) lattice,[2] and therefore would not exhibit this problem.

3. $P$ and $\neg P$ are essentially treated independently. The certainty of $P$ is represented by the LB of $P$. The certainty of $\neg P$ is represented as the negation of the UB of $P$. With one exception, the LB and UB do not affect each other. That exception is when the LB becomes greater than the UB, which roughly corresponds to the situation when both $P$ and $\neg P$ are believed with high degrees of certainties. When this happens a conflict handler tries to detect the source of the inconsistency.

4. Disjunctions in the conclusions of rules are not allowed.

These restrictions amount to allowing only acyclic quantitative Horn clauses. The following definitions formalize some of the above constraints.

**Definitions:** A RUM specification is a triple (W, I, J). W is a set of wff's, such that whenever w ∈ W, $\overline{w}$ ∈ W. For w ∈ W, LB(w) ∈ [0, 1] is the amount of evidence confirming the truth of w. 1 − UB(w) ∈ [0, 1] is the amount of evidence confirming the falsity of

---

[1] This boundary conditions resolve the possible indeterminations of the T-norms $T_i$ and T-conorms $S_i$ defined in the paper, i.e.,

$$T_i(0,x) = T_i(x,0) = 0, \quad S_i(1,x) = S_i(x,1) = 1 \quad \forall x \in [0,1]$$

[2] Idempotency (or equivalently, Non-Archimedean) is the the required T-norm property.

w. $LB(\overline{w}) = 1 - UB(w)$.[3] $I \subset W$, is a distinguished set of input wffs, that could potentially take on values from the outside world. J is a set of justifications. Each justification is a triple (P, s, c). $P \subset W$ is the premises of the justification. $s \in [0, 1]$ is the sufficiency of the justification. $c \in W$ is the conclusion of the justification.

**Definition:** A conflict occurs when $\exists\ w \in W$ s.t. $LB(w) + LB(\overline{w}) > 1$.

**Definitions:** A RUM rule graph is a triple (A, O, E). Where A is the AND nodes in an AND/OR graph, O is the OR nodes, and E is the arcs (n1, n2). The RUM rule graph (A, O, E) of a RUM specification (W, J) is given by A = J, W = O, and $(j \in J, w \in W) \in E$ iff $J=(P, s, c) \wedge w = c$. $(w \in W, j \in J) \in E$ iff $J=(P, s, c) \wedge w \in P$. Additionally each arc emanating from a justification is labeled with a real number $\in [0, 1]$ representing its contribution to its conclusion.

**Definition:** A valid RUM specification is one in which the corresponding rule graph is acyclic.

**Definition:** A RUM rule graph is admissible iff:

1. the label of each arc leaving a justification equals the T-norm of the arcs entering the justification and the LBs of the premises of the justification and

2. the LB of each wff is the S-conorm of the labels of the arcs entering it.

Due to these restrictions a simple linear time algorithm is able to propagate the correct numeric bounds through a valid RUM rule graph to generate an admissible rule graph. The only possible exponential step is when the conflict handler has to resolve an inconsistency.

### 3.3 PRIMO's Extensions to RUM

RUM deals with missing information in a *procedural* form, by attaching *demons* to the ignorance measure of the value assignments to a variable. RUM, however, does not provide any *declarative* representation to handle incomplete information.

In a recently proposed approach [BCGS89], we have addressed the problem integrating defeasible reasoning, based on non-monotonically justified rules,[4] with plausible reasoning, based on monotonic rules with partial degrees of sufficiency and necessity.

In this approach, uncertainty measures are propagated through a directed acyclic graph (DAG), whose nodes can either be object-level wffs or non-monotonic loops. The links in the DAG are plausible inference rules with Horn clause restrictions. The non-monotonic loops are composed of non-monotonically justified rules. The key idea is to exploit the information of the monotonic links carrying uncertainty measures, by creating a preference function that will be used to select the extension, i.e., the fixed point of the non-monotonic loop, which is maximally consistent with the soft constraints imposed by the monotonic links. Thus, instead of minimizing the cardinality of abnormality types [McC86] or of performing temporal minimizations [Sho86], we maximize an *information-content* function based on the uncertainty measure. This method breaks the symmetry of the (potentially) multiple fixed points in each loop by selecting the most likely extension. This idea is currently being implemented in PRIMO (Plausible ReasonIng MOdule), RUM's successor.

## 4 The Integrated RUM/RUM-runner Technology

### 4.1 The Rapid Prototyping Paradigm

We have observed that dynamic classification problems are characterized by an evolving set of requirements and need the use of the rapid prototyping methodology for their development [Pre87]. The prototypes are developed in rich and flexible environments in which various AI techniques are used. A knowledge base is generated, debugged, modified, and tested until a "satisficing" solution is obtained from this development phase. Then the prototype is ready for deployment: it is ported to specific platforms and embedded into larger systems. The deployment's success, however, depends on the application performing in *real-time*. If the reasoning system does not provide good timely information, then the application will not be able to react fast enough to its environment. Even after deployment, the prototype cycle must continue, because performance verification can only take place in a real-time environment. Thus, in order to meet the real-time requirements, the knowledge base and algorithms may need additional prototyping.

AI software development is significantly different

---

[3] Note this is the same relationship as that between support and plausibility in Dempster-Shafer theory and that between □ and ◇ in modal logics.

[4] A non-monotonically justified rule $j$ is of the form:

$$\bigwedge_i ma_i \wedge \bigwedge_i nma_i \rightarrow^s c$$

where $s \in [0, 1]$, the *sufficiency* of the justification, indicates the confidence of the justification; $ma_i \in L$, are the monotonic antecedents of $j$; $nma_i$ are the nonmonotonic antecedents of $j$, and have the form, $\neg \boxed{\alpha} p$, where $p \in L$, with the semantics:

$$LB(\neg \boxed{\alpha} p) = \begin{cases} 0 & \text{if } LB(p) \geq \alpha \\ 1 & \text{if } LB(p) < \alpha \end{cases}$$

26

from the traditional approach. It requires a prototyping cycle which spans two environments: development and target. Usually, instead of having to transition software between these two environments, one environment is eliminated. This approach, however, compromises either the flexibility and richness needed for development, or the speed and efficiency requirements of execution. When both environments are used, a smooth transition of the application between these two environments is essential. If the prototyping cycle cannot completely span the two environments, the knowledge engineer has to re-implement portions of the software.

The development of reasoning systems addressing dynamic classification problems presents another difficulty: *testing and validating the knowledge base and inference techniques*. To solve this problem we have implemented an integrated software architecture capable of generating, interpreting, and resolving complex time-varying scenarios. The test-bed architecture is composed of two parts: a *simulation environment* capable of maintaining the dynamic states of numerous simulated objects; and a *reasoning system* capable of dealing with the uncertain, incomplete, and time-varying information.

### 4.2 Simulation Environment

The simulation environment is composed of four basic modules: the *window subsystem*, a window based user interface for displaying maps; the *annotation subsystem*, an intelligent database for displaying time varying features; *LOTTA*, the simulator; and a set of tools for interfacing to a reasoning system. LOTTA is a symbolic simulator implemented in an object-oriented language (Symbolics Flavors). LOTTA maintains time varying situations in a multiple player antagonistic game where players assess situations and make decisions in light of uncertain and incomplete data. LOTTA has no reasoning capabilities; these are provided by external reasoning modules, easily interfaced to the LOTTA data structures. LOTTA is further described in [BA88].

### 4.3 Reasoning System: RUM

The integrated reasoning system is composed of RUM [BGD87], a rich, user-friendly development environment, and RUMrunner, a small and quick run-time system, and translation software to span the two (see Figure 2).

RUM embodies the theory of plausible reasoning described in the previous section. RUM provides a representation of uncertain information, uncertainty calculi for inferencing, and selection of calculi for inference control. Uncertainty is represented in both facts and rules. A fact represents the assignment of a value to a variable. A rule represents the deduction of a new fact (conclusion) from a set of given facts (premises). Facts are qualified by a degree of *confirmation* and a degree of *refutation*. Rules are discounted by *sufficiency*, indicating the strength with which the premise implies the conclusion, and *necessity*, indicating the degree to which a failed premise implies a negated conclusion. The uncertainty present in this deductive process leads to considering several possible values for the same variable. Each value assignment is qualified by different uncertainties, which are combined with T-norm based calculi as described in [Bon87b] and [Bon87c].

RUM's rule-based system integrates both procedural and declarative knowledge in its representation. This integration is essential to solve situation assessment problems, which involve both heuristic and procedural knowledge.

The expressiveness of RUM is further enhanced by two other functionalities: the *context mechanism*, to determine the rule's applicability to a given situation, and the *belief revision*, to detect changes in the input, keep track of the dependency of intermediate and final conclusions on these inputs, and maintain the validity of these inferences. These features will be described in the next section, to illustrate their role in solving the KB scalability issue. RUM is further elaborated upon in [BGD87].

These AI capabilities are used to develop a knowledge base, in conjunction with RUM's software engineering facilities, such as flexible editing, error checking, and debugging. Some of these features, however, are no longer necessary once the development cycle is complete. At run-time, applications do not create new knowledge (facts or rules), as their basic structures have been determined at compile-time. The only run-time requirement is the ability to instantiate rules and facts from their pre-determined definitions. By eliminating the development features which are unnecessary at run-time, a real-time AI system can improve upon the algorithms and methodologies used in RUM.

### 4.4 Reasoning System: RUMrunner

The objective of RUMrunner [Pfa87] is to provide a software tool that transforms the customized knowledge base generated during the development phase, into a fast and efficient real-time application. RUMrunner provides both the functionality to reason about a broad set of problems, and the speed required to properly use the results of the reasoning process. Performance improvements are obtained by implementing all RUM's functionalities with leaner data structures, using Flavors (for the Symbolics version) or *defstructs* (for the Sun version). Furthermore, RUMrunner no longer requires the use of the KEE software, thus it can be run on any Symbolics or Sun workstation with much smaller memory configurations, and without a KEE software license. RUMrunner has four major qualities: it provides a meaningful subset of AI techniques, it runs fast, it has the functionality of a real-time system, and it does not require the software engineer to re-program the application in the target environment.

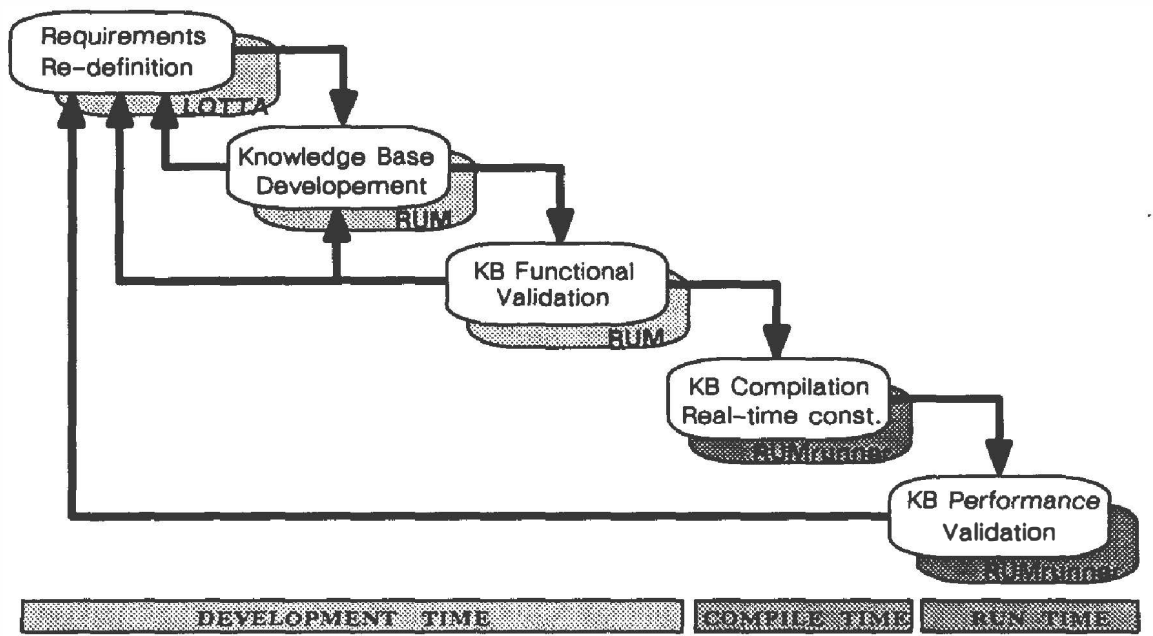

Figure 2: Software Engineering with RUM and RUMrunner

To increase speed, RUMrunner takes advantage of the fact that the application has been completely developed and debugged. It provides a minimum of error checking because the application is assumed either to be debugged already, or to be robust enough to handle errors. RUMrunner's time performance in reasoning tasks is partially due to the compilation of the knowledge base. As a result of this compilation, new or different rules or units cannot be created in the knowledge base after the translation.

RUMrunner provides additional functionality for applications which must satisfy real-time requirements. A RUMrunner application is able to carry out and control a set of activities to rapidly respond to its environment. To meet these goals, the interface of RUMrunner with the application program is designed to be asynchronous, allowing the application to avoid unnecessary delays. In addition, the application is able to handle externally or internally driven interrupts. It is also able to prioritize tasks, by using an agenda mechanism, so that RUMrunner handles the most important ones first. RUMrunner is performance-conscious by ensuring that tasks execute within a specified amount of time. This is done through planning the execution of a single task as suggested by Durfee and Lesser [DL87]. Finally, RUMrunner is implemented in Common LISP, thus it can be ported to many machines without requiring any proprietary software. RUMrunner, is further elaborated upon in [Pfa87].

The approach followed in developing RUMrunner is predicated on the following three steps: *Knowledge Base compilation*; *execution-time estimation*; *run-time/real-time execution*. This process is indicated in Figure 3.

### 4.4.1 Knowledge Base Compilation

One of the most natural steps used to improve the performance of a program is to compile it to avoid unnecessary run-time checks, searches, and value substitutions. In the case of a knowledge base, the compilation is done at the representation language level (beside the traditional compilation from source code to object code, done at the programming language level). Following this philosophy, we have developed a RUM-rule compile, composed of three components:

1. *Reader/Translator* - to read the information stored in the hierarchical rule base, originally written in customized macro-expressions and then expanded into a graph of KEE Units

2. *Analyzer* - to derive and analyze the rule topology (identifying the sharing of common structures), and to create the pointers between the

28

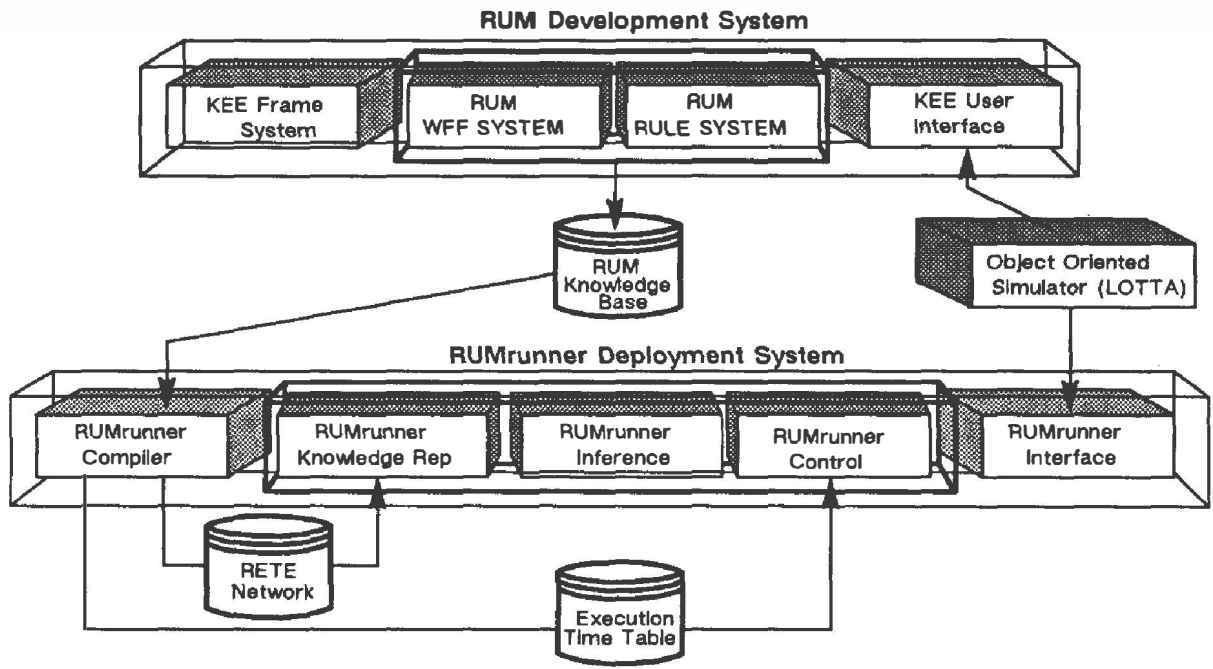

Figure 3: Uncertainty Tool Set Architecture

declarative part of the knowledge base (captured by the rule dependency) and the procedural part of the knowledge base (captured by the system or user-defined predicates)

3. *Code Generator* - to write the directed acyclic graph, produced by the Analyzer as a modified RETE network, into an output file to be used by the run-time/real-time inference engine.

### 4.4.2 Execution-time Estimation

During the development of the knowledge base, an object oriented simulator (LOTTA) was used to create and run a variety of scenarios representing a substantial set of requirements to be met by the reasoning system. The simulation runs generated numerous track-files, which were then stored. After the rule compilation, the same representative data from the track-files is used to fire each individual predicate and rule. During this process, their execution times are measured and logged into a dedicated table of performance timing. This table is then used at run-time by the planner to determine the cost (in terms of execution time) of using deductive paths which require the re-firing of specific predicates and rules.

### 4.4.3 Run-time/Real-time Execution

The real-time execution of RUMrunner is supported by the following features:

**Automatic Rule Instantiation.** Rule template, written by the knowledge engineer, contain variables scoped by a universal quantifier. This quantifier is replaced by an enumeration of its instances, as new objects are created in the scenario, and their properties bound to the rule slots. The rule templates are automatically instantiated as new objects (linked to the rules) are created during the scenarios.

**Caching of Node State.** Each node in the Direct Acyclic Graph (DAG) generated by the rule compiler, is either a *variable*, a *predicate*, or a *rule*. Each node has its own state: variable-nodes store all their current value-assignments (and their tests on the value distribution of the tested variables; rule-nodes store the results of their detachments (rule conclusions), which are asserting a value assignment for some other variable-node in the network. The local state of each node in the DAG has a validity flag, indicating whether the information in the cache can be fetched, or should be recomputed due to changes or obsolescence. The validity flag are maintained by a belief revision system similar to the one used in RUM.

**Asynchronous Processing.** An agenda mechanism is used to asynchronously receive any number of input tasks (such as backward-chaining on a goal or forward-chaining on a given piece of evidence) from various sources. The agenda scheduler (see below) sorts the tasks according to their characteristics. The outputs of RUMrunner are isolated from other connecting systems via buffers or streams.

**Task Priorities and Deadlines.** Each task in the agenda receives a (static) level number, determining the relative priority of the task with respect to the other ones. A time-deadline, expressed in absolute time, is attached to the task to indicate its urgency (i.e., its expiration time). The task are sorted by priority and, within the same priority level, by the shortest deadline.

**Individual Processes.** Each task in the agenda is assigned a separate process. Such process is then terminated, once the task is completed. This capability requires an underlying operating system supporting multi-tasking.

**Interrupts.** External or internal interrupts, with re-entrant reasoning, can superseed the current task. Three types of interrupts have been implemented: the internal interrupts caused by queries approaching their assigned time-deadlines; the external interrupts caused by queries with higher priority than the one currently addressed; and the external interrupts caused by new input data characterized by higher priority than the current query.

**Scoping the KB by Rule Classes.** Design-time partitions of the Knowledge Base, expressed by a hierarchy of rule classes and sub-classes, can be exploited at run-time by adding to the task an optional argument describing the subset of the KB (denoted as a list of rule classes/sub-classes), which is relevant to the task. As a result, forward and backward chaining can be scoped by this optional argument.

**Planning to Meet Deadlines.** A run-time planning system determines the largest amount of redundancy (parallel proof trees) which the system can afford to use and still meet its time-deadline. At run-time, a graph traversal determines the validity flags of the nodes in the sub-graph used to solve the task. The cost of those nodes whose validity flag is requiring a re-execution are retrieved from the tables. With this information the total estimated time for the path is computed. By taking into account the sharing of common nodes in different paths, the planner maximizes the coverage of the sub-graph that can be executed within the allocated time budget (determined by the task deadline and the current clock-time). A current option for the planner is to use an upper bound of the amount of certainty potentially provided by each path in the compiled network (i.e., the minimum of the sufficiency values attached to the links of the path). This bound is a static measure of information content, which can be used in conjunction with the estimated execution cost to determine the set of reasoning paths to be used.

# 5 Addressing the DCP's Reasoning Requirements

## 5.1 Addressing the Representation, Inference, and Control Issues

We will now illustrate how the RUM/RUMrunner technology can be used to meet the requirements described in Section 2.

1. RUM has a consistent and efficient way to represent and propagate uncertainty through the rule base. RUM handles uncertain information by qualifying each possible value assignment to any given variable with an uncertainty interval. The interval's lower bound represents the minimal degree of confirmation for the value assignment. The upper bound represents the degree to which the evidence failed to refute the value assignment. The interval's length represents the amount of ignorance attached to the value assignment. The uncertainty intervals are propagated and aggregated by five Triangular norm based uncertainty calculi. The T-norms' associativity and truth functionality entail problem decomposition and relatively inexpensive belief revision. This characteristics makes RUM an intensional system [Pea88] with a *modular* inference layer, based on *locality* and *detachment*.

2. RUM represents defaults (i.e., decision maker's preferences or biases) by means of active values, which monitor the amount of information content (i.e., ignorance) of a given slot and assert the defaults when no plausible evidence or conclusion is available.

3. Such default assertions are retracted by the system, when a reliable evidence or conclusion is encountered by the system.

4. RUM's belief revision is essential to the dynamic aspect of the classification problem. The belief revision mechanism detects changes in the input, keeps track of the dependency of intermediate and final conclusions on these inputs, and maintains the validity of these inferences. For any conclusion made by a rule, the mechanism monitors the changes in the certainty measures that constitute the conclusion's support. Validity flags are used to reflect the state of the certainty. For example, a flag can indicate that the uncertainty measure is valid, unreliable (because of a change in the support), too ignorant to be useful, or inconsistent with respect to the other evidence.

5. Following the testing, and verification of the application using RUM, the knowledge base gen-

erated by RUM is then automatically translated and compiled into compact data structures. RUMrunner reasons opportunistically with these data structures to achieve the run-time performance required by most real-time applications.

## 5.2 Addressing the Scalability Issue

A crucial aspect of a good architecture is the ease with which it can be *functionally decomposed*. Using the RUM/RUMrunner technology, we can develop different modules as separate knowledge bases, perform local functional testing for each module, interconnect the modules, and test the entire system. Each module (Knowledge Base) can be *statically* and *dynamically* partitioned. At design time, a module is statically subdivided into a hierarchy of rule-classes/subclasses, which reflects the existing functional decomposition within the module. Rule classes/sub-classes can be used by RUM/RUMrunner to determine the scope of the forward/backward chaining, with the purpose of limiting the number of rules evaluated at run-time. This locality effect minimizes unwanted rule interactions, and improves run-time efficiency.

The RUM/RUMrunner tool also provides the knowledge engineer with a dynamic partition mechanism, referred to as *context mechanism*. The context represents the set of *preconditions* determining the rule's applicability to a given situation. These preconditions can be described by predicates on object-level variables (such as the quadrant location of an oncoming object) or by predicates on meta-level variables (such as restrictions on the types of sensor to be used or the amount of available time before an action is due). This mechanism provides an efficient dynamic screening of the knowledge base by focusing the inference process on small rule subsets.

These (static and dynamic) partitions represent an approximation to an optimal solution derived by using the entire knowledge base. Theoretically, they could cause a potential loss of correctness and completeness in the solution. Pragmatically, they are essential to ensure proper functional testing, expandability, and maintenance of the application software.

## 5.3 Addressing the Software Engineering Problems

Figure 2, in the previous section, illustrates the cascading tasks associated with the development of a knowledge base application. The first three tasks (*Requirement Re-definition, KB Development, KB Functional Validation*) are performed in the development environment. The last two tasks (*Product Engineering and Performance Validation*) are performed in the deployment system. We will focus on the validation tasks in both development and deployment systems.

### 5.3.1 Addressing the Functional Validation Issue

The objective of this task is to assure that the knowledge base will meet the requirements derived from the problem definition. We have used LOTTA to generate a set of scenarios (sequence of events), which collectively exercise all the desired requirements. For instance, these scenarios have allowed us to test the robustness of the KB in light of unexpected events, such as the appearance of a second platform in a one-on-one situation, or while reasoning with reduced information to reflect constraints on the use of the ownship's active sensors, etc. By interactively modifying LOTTA's scenarios, we have tested the reasoning system on a class of scenarios with multiple variations, representing "what-if" type of situations.

In all these scenarios, LOTTA maintained ground truth (i.e., states and sets of orders of all the players' objects.) At the end of each sensor phase, LOTTA generated the corresponding track file information representing the perceived truth of the simulated world. These track files have then been used to test the rule set for consistency and completeness. The same track files, stored as buffers, have later been applied as probing input to exercise the run-time system.

RUM's conclusion's explanation and traceability facilities have been used to identify and analyze the dominant rules responsible for specific conclusions. By comparing the conclusions with ground truth, the knowledge engineer has been able to detect and correct eventual discrepancies. This corrective process was achieved by verifying the validity of the input to the rule set (track file information), by examining the context of the active rules, by analyzing the structure of the active rules (under or over constrained), by calibrating the strength of the dominant rules (sufficiency and necessity), and by modifying the sensitivity to uncertainty exhibited by the dominant rules (uncertainty calculus selection).

### 5.3.2 Addressing the Performance Validation Issue

The objective of this task is to guarantee that the software will meet the timing requirements imposed by the real-time constraints, while still maintaining the same functional behavior.

As described in the above sections, this goal is achieved by a combination of efforts: the translation of RUM's complex data structure into simpler, more efficient ones (to reduce overhead); the compilation of the rule set into a modified RETE net [Mir87] (to avoid run-time search); the load-time estimation of each rule's execution cost (to determine, at run-time, the execution cost of any given deductive path); the run-time planning mechanism for model selection (to determine the largest relevant rule subset which could be executed within a given time-budget). In some of the application functions we have experienced performance improvements of more than 200 times with respect to the original RUM application.

### 5.3.3 Addressing the Software Portability Issue

Currently, RUM runs on top of KEE on Sun Workstations and Symbolics. RUMrunner runs on SUN workstations with Lucid Common LISP and Symbolics. To increase the number of potential delivery platforms for RUMrunner, we are modifying the RUM's Knowledge Base Compiler and we are rewriting RUMrunner inference engine. The Code Generator sub-module, in RUM's Knowledge Base Compiler, writes the directed acyclic graph, produced by the Analyzer as a modified RETE network, into an output file to be used by the run-time/real-time inference engine. By modifying the Code Generator sub-module, we can write an Ada or C output file, containing the same modified RETE network.

Finally by rewriting the run-time inference engine in the same programming language (Ada or C, respectively), we can interpret the knowledge compiled into the RETE network and use it with the run-time data. Thus, all the declarative knowledge, contained in the rule base and encoded into the RETE network, can be cross-compiled to run on different platforms, using different programming languages. The procedural knowledge used by the application, encoded in the predicate file used by RUMrunner, will have to be manually translated to other programming languages. Usually, the size of this file will be relatively small, as predicates are shared by many rules.

## 6 RUM/RUMrunner Applications

Over the last two years, this technology has been applied to a variety of dynamic classification problems, a subset of which are illustrated in this section.

The first application of this technology has been the Situation Assessment (SA) Module of DARPA's Pilot's Associate (PA) Program. Pilot's Associate aims to improve the combat effectiveness of post-1995 aircraft through the application of mature AI technologies. The Situation Assessment Module is one of six modules within the overall PA system. SA is concerned with analyzing the external environment, such as determining enemy threats to ownship. RUM has been used by GE-CRD and Lockheed Georgia to develop two submodules within the Situation Assessment module. One submodule determines the class and type of enemy aircraft; the second submodule determines the Target Value of an aircraft. Target Value is composed of target importance, target opportunity, target capability and the intent of the bogey. The total KB size is approximately 200 rule-templates.

Another situation assessment module, applied to the submarine problem domain, is currently been developed by the GE-CRD AI Program. The actual Knowledge Base, developed over the last three months, contains about 300 rule-templates to determine contact's knowledge of ownship (counter-detection, location of ownship and threat alert), threat maneuvers of contacts (straight-line drive, turn-away-and-run, turn-to-and-reengage, secondary attack) and the intent of the contact (attack, evade, influence, non-reactive).

Other internal RUM applications include the navigation system for the Mars Roving Vehicle and in-flight diagnosis of avionics. All these applications have improved our understanding of the Dynamic Classification Problems and have provided us with valuable feed-back and guidelines to direct our efforts in refining the RUM/RUMrunner technology.

## 7 Conclusions

We have defined a theory for reasoning with uncertainty based on the semantics of many-valued logics (T-norm operators). We have implemented a subset of this theory, limited to acyclic Horn clauses, and embedded in a software tool. We have integrated this tool with other technologies (object-based simulators, real-time inference engines) into a software architecture designed to address the class of Dynamic Classification Problems. Finally, we have applied this software architecture to a variety of DCP cases: situation assessment (for Pilot's Associate and for Submarine Commander's Associate), navigation and diagnosis.